\documentclass[preprint,aps]{revtex4}

\usepackage{graphicx}
\usepackage{dcolumn}
\usepackage{bm}

\begin{document}

\preprint{APS/123-QED}

\title{Evolving Stochastic Learning Algorithm Based on Tsallis Entropic Index} 
\author{Aristoklis D. Anastasiadis}
 \email{aris@dcs.bbk.ac.uk}
\altaffiliation[AD is also affiliated with]{London Knowledge Lab, University of
London, 23-29 Emerald Street, WC1N 3QS, London, United Kingdom
}
\author{George D. Magoulas}%
 \email{gmagoulas@dcs.bbk.ac.uk}
\affiliation{%
School of Computer Science and Information Systems, Birkbeck
College,
 University of London, Malet Street, London  WC1E 7HX, United Kingdom.\\
}%


\date{\today}

\begin{abstract}
In this paper, inspired from our previous algorithm, which was
based on the theory of Tsallis statistical mechanics, we develop a
new evolving stochastic learning algorithm for neural networks.
The new algorithm combines deterministic and stochastic search
steps by employing a different adaptive stepsize for each network
weight, and applies a form of noise that is characterized by the
nonextensive entropic index $q$, regulated by a weight decay term.
The behavior of the learning algorithm can be made more stochastic
or deterministic depending on the trade off between the
temperature $T$ and the $q$ values. This is achieved by
introducing a formula that defines a time--dependent relationship
between these two important learning parameters. Our experimental
study verifies that there are indeed improvements in the
convergence speed of this new evolving stochastic learning
algorithm, which makes learning faster than using the original
Hybrid Learning Scheme (HLS). In addition, experiments are
conducted to explore the influence of the entropic index $q$ and
temperature $T$ on the convergence speed and stability of the
proposed method.
\end{abstract}


\maketitle

\section{\label{sec:level1}Introduction}\vspace*{-0.25cm}
Neural networks are widely used in many classification
applications. One of the major key concept in neural networks is
the interaction between microscopic and macroscopic phenomena. The
goal of Feedforward Neural Network (FNN) learning is to
iteratively adjust the weights, in order to globally minimize a
measure of the difference between the actual output of the network
and the desired output, as specified by a teacher, for all
examples ($P$) in a training set~\cite{Haykin94}:
\vspace*{-0.25cm}
\begin{footnotesize}
\begin{equation} \label{error_E}
E(w)={\sum_{p=1}^P} \, {\sum_{j=1}^{n_L}} {{\left({y_{j,p}^L} -
t_{j,p}\right)}^2} \\
= {\sum_{p=1}^P} \, {\sum_{j=1}^{n_L}} {{\left[{
{\sigma}^L\kern-3pt\left( \, net_j^L+ {{\theta}^L_j}\right)}-
t_{j,p}\right]}^2}.
\end{equation}
\end{footnotesize}
\noindent where, $net_j^L$ is for the $j$-th node in the $l$-th
layer ($j=1,\dots,n_L$), the sum of its weighted inputs.
${\theta}^L_j$ denotes the bias of the $j$--th node ($j=1, \dots,
N_l$) at the $l$--th layer ($l=2, \dots, L$), and $w$ denotes the
weights $w$ in the network. This equation formulates the energy
function, called {\em error function}, to be minimized, in which
$t_{j,p}$ specifies the desired response at the $j$--th output
node for the example $p$ and ${y^L_{j,p}}$ is the output of the
$j$--th node at layer $L$ that depends on the weights $w$ of the
network, and ${\sigma}$ is a nonlinear activation function, such
as the well known logistic function $\sigma(x) =
{(1+e^{-x})}^{-1}$. The problem of finding the global minimum of
such a complex cost function, which possesses a large number of
local minima, is considered very difficult task~\cite{Haykin94}.

Statistical mechanical methods have  been applied successfully to
the study of neural network models of associative
memory~\cite{Gyorgyi01}. These models are biologically plausible
and can be trained very quickly in some cases, compared with the
popular neural networks such as multi--layered perceptron, which
have been shown to work satisfactorily. However, this model of
associative memory has still drawbacks as learning gets stuck at
local minima. A variety of global optimization algorithms have
also been introduced over the years to overcome the problem of
local minima. One of the most popular methods is the Simulated
annealing~\cite{KirkpatrickGV83}. It uses Boltzmann--Gibbs (BG)
statistics at two different steps, namely at the \emph{visitation
step}, which uses a Gaussian distribution, and at the
\emph{acceptance step}, that uses the Boltzmann
factor~\cite{AckleyHS85,ArtsK89}.

Another approach is based on the use of noise models. Attempts to
explore the benefits of introducing noise during learning have
been based on the use of Gaussian
distributions\cite{AckleyHS85,BurtonM92,Rognvaldsson94}. One of
the most famous neural model operating with noise is the Boltzmann
machine,~\cite{AckleyHS85,ArtsK89}, inspired by the
Boltzman--Gibbs entropy $S_{BG}=-K\sum_{i}p_{i}ln p_{i}$ that
provides exponential laws for describing stationary states and
basic time--dependent phenomena, where $\{p_i\}$ are the
probabilities of the microscopic configurations, and $K>0$. Also,
a form of Langevin noise has been proved quite effective for
neural learning, and has motivated the development of other
methods, such as the {\em Simulated Annealing
Rprop}--SARprop~\cite{TreadgoldG98}.

The next section briefly describes the recently proposed hybrid
learning scheme~\cite{AnastasiadisM2004a}, and then we introduce
the proposed evolving stochastic learning algorithm. Next, results
of an empirical evaluation are presented, demonstrating the
effectiveness of the new scheme in locating acceptable solutions.
The paper ends with discussion and concluding remarks.
\vspace*{-0.5cm}
\section{\label{sec:level2}The Evolving Stochastic Learning Algorithm}
\vspace*{-0.25cm} The recently proposed Hybrid Learning Scheme
(HLS)~\cite{AnastasiadisM2004a} has been built on ideas from
global search methods. It is worth noting that global search
algorithms possess strong convergence properties. However, these
methods are computationally expensive~\cite{TreadgoldG98}. To
alleviate this situation hybrid schemes for neural networks
learning have been developed in an attempt to achieve improved
convergence rates compared to the standard global optimization,
and in some cases even maintain the guarantee of convergence to a
global minimizer~\cite{BurtonM92}. HLS is a hybrid training
algorithm that employs a different adaptive stepsize for each
weight. HLS avoids slow convergence in the flat directions and
oscillations in the steep directions, and exploits the parallelism
inherent in the evaluation of learning error $E(w)$ and gradient
$\nabla E(w)$ by the Resilient Back-Propagation (Rprop)
algorithm~\cite{RiedmillerB93}. Inspired
by~\cite{BurtonM92,TsallisS96}, in the HLS, noise has been
introduced in the training procedure according to a nonextensive
schedule~\cite{AnastasiadisM2004a}. The HLS also applies the
sign--based weight adjustment of Rprop~\cite{RiedmillerB93}, on
the perturbed energy function (for a detailed description see
~\cite{AnastasiadisM2004a}).

The new Evolving Stochastic Learning Algorithm~(ESLA) introduces
noise, as in HLS. The noise source is characterized by the
nonextensive entropic index $q$. In particular, the principles of
the new method are using the notion of nonextensive entropy, which
has been defined as~\cite{Tsallis88}: \vspace*{-0.25cm}
\begin{equation}\label{eq:tsallis Sq}
    S_{q}\equiv K\; \frac{1-\sum_{i=1}^{W}p_{i}^{q}}{q-1} \;\;\;\;
    (q \in {R}),
\end{equation}
\noindent where $W$ is the total number of microscopic
configurations, whose probabilities are $\{p_i\}$, and $K$ is a
conventional positive constant. When the entropic index $q=1$,
~(\ref{eq:tsallis Sq}) recovers to Boltzmann--Gibbs entropy. The
entropic index works like a biasing parameter: $q<1$ privileges
rare events (values of $p$ close to 0 are benefited), while $q>1$
privileges common events (values of $p$ close to 1). The
optimization of the entropic form~(\ref{eq:tsallis Sq}) under
appropriate constraints,~\cite{Tsallis88}, yields for the
canonical ensemble
\vspace*{-0.25cm}
\begin{equation}\label{eq:tsallis pi}
    p_i \propto [1-(1-q) \beta E_i]^{\frac{1}{(1-q)}} \equiv e_{q}^{-\beta E_i},
\end{equation}
\noindent where $\beta$ is a Lagrange parameter, $\{E_i\}$ is the
energy spectrum, and the $q$-{\em exponential function}
\vspace*{-0.25cm}
\begin{equation}\label{eq:tsallis q-exp}
    e_{q}^{x}\equiv [1+(1-q) x]^{\frac{1}{(1-q)}}=\frac{1}{[1-(q-1)x]^{\frac{1}{(q-1)}}}
\end{equation}
\noindent In this method, like in the HLS, noise is generated
according to a schedule: \vspace*{-0.25cm}
\begin{equation}\label{sa term}
    Q (T,k)= e_{q}^{-T(\ln2)\cdot k} = [1-(1-q)T(\ln2)\cdot k]^{\frac{1}{1-q}},
\end{equation}
\noindent where $T$ is the temperature; $k$ indicates iterations.
Noise is not applied proportionally to the size of each weight;
instead a form of weight decay is used, which is considered
beneficial for achieving a robust neural network that generalizes
well. Thus, noise is introduced by formulating the {\em perturbed}
energy function:
\vspace*{-0.25cm}
\begin{equation} \label{noisy_error_E}
\tilde{E}(w^k) = E(w^k)+ \mu \cdot {\sum_{i=1}^n {{(w_i^k)}^2
\over [1+{(w_i^k)}^2]}}\cdot Q(T,k),
\end{equation}
\noindent where $E(w)$ is the error function, ${\sum_i {w_i^2 /
(1+w_i^2)}}$ is the weight decay bias term which can decay small
weights more rapidly than large weights, and $\mu$ is a parameter
that regulates the influence of the combined weight decay/noise
effect. The energy landscape is modified during training so the
search method is allowed to explore regions of the energy surface
that were previously unavailable. Minimization
of~(\ref{noisy_error_E}) requires calculating the gradient of the
energy with respect to each weight \vspace*{-0.25cm}
\begin{equation} \label{noisy_grad}
\tilde{g}_i(w^{k}) = {g}_i(w^{k}) + \mu\acute{} \cdot {{w_i^k}
\over {{[ 1 + {(w_i^k)}^2 ]}^2}}\cdot Q(T,k),
\end{equation}
\noindent where ${g}_i(w^{k})$ is the gradient of the energy
$E(w^k)$, with respect to each weight, and $\mu\acute{}>0$ (in our
 experiments a fixed value of
$\mu\acute{}=0.01$ was used). The proposed evolving stochastic
hybrid scheme applies a sign--based weight adjustment, similar to
HLS~\cite{AnastasiadisM2004a}, on the perturbed energy
function~(\ref{noisy_error_E}) using the gradient term of
Equation~(\ref{noisy_grad}). Also the learning rates are adapted
by Rprop learning procedure~\cite{RiedmillerB93}.

In our approach the weight adjustment is given by the following
equation:
 \begin{equation}\label{GRprop}
       w^{k+1} = w^k - \tau^k\, {\rm diag}\{\eta_1^{k}, \ldots,\eta_i^{k},\ldots,
       \eta_{n}^{k}\}\,{\rm sign} (\tilde{g}_i(w^{k})),\;\;\;\;\;\;k=0,1,\ldots
 \end{equation}
where ${\rm sign} (\tilde{g}_i(w^{k}))$ denotes the column vector
of the signs of the components of
$\tilde{g}(w^k)=\left(\tilde{g}_1(w^k), \tilde{g}_2(w^k), \ldots,
\tilde{g}_n(w^k)\right)$, $\tau^k > 0$, $\eta_m^{k}$
($m=1,2,\ldots,i-1,i+1,\ldots,n$) are small positive real numbers
generated by Rprop's learning rates schedule.

Moreover, an additional condition, like in the HLS, is introduced
in order to avoid using relatively small weight adjustments
\vspace*{-0.25cm}
\begin{eqnarray}
&&\!\!\!\!\!\!\!\!\!\!\!\!\!
if \quad \left( \eta_{i}^{k-1} < \rho \cdot Q^2(T,k) \right) \quad then \nonumber\\
&&\!\!\!\!\!\!\!\!\!\!\!\!\!\! \quad \eta_{i}^k = max\left(
\eta_{i}^{k-1}\/ \eta^{-} + 2 c \rho \cdot Q^2(T,k),
\Delta_{min}\right), \label{RpropETA4}
\end{eqnarray}
\noindent where $0<\rho<1$ and $c \in (0, 1)$ is a random number.

Lastly, inspired from previous work,~\cite{TsallisS96}, we apply a
cooling procedure. This defines the relationship between $T$ and
$q$ values. The application of cooling helps to regulate the
training algorithm, making it more deterministic. This new {\em
Evolving Stochastic Learning Algorithm}-ESLA behaves in a more
stochastic way, during the initial stages, and then becomes more
deterministic as the number of iterations increases. Thus, when we
are close to the minimizer, the algorithm hopefully will avoid
oscillations and converge faster. The cooling procedure is
described by the next equation:
\vspace*{-0.25cm}\begin{equation}
\label{Cooling Temp}
T=T_{0}\cdot[\frac{2^{q-1}-1}{(1+k)^{q-1}-1}],q>1
\end{equation}
where $T_{0}$ is the initial temperature, $T$ is the current
temperature, $k$ is the number of iterations, and $\textit{q}$ is
the Tsallis entropic index.

The challenge is to cool the temperature the quickest we can, but
still having the ability to converge to global minimum with high
probability. The standard simulated annealing (SA) is one method
to achieve this goal. However, the cooling procedure is
computationally expensive. An efficient alternative cooling method
is the fast simulated annealing (FSA)~\cite{Szu87}. The
temperature is now allowed to decrease like the inverse of time,
which makes the entire cooling procedure quite more efficient.
Simulated annealing (GSA)~\cite{TsallisS96} is a generalization of
the previous methods, which performs better than previous
annealing algorithms for many problems and applications. In neural
networks applications we are mainly interested in accelerating the
learning speed with no affect in generalization. The cooling
procedure based on GSA satisfies these two targets and contributes
positively to the performance of the ESLA. This cooling procedure
makes the temperature to decrease as a power-law of time, in
contrast to the much slower decrease (logarithmic in time) of the
$q=1$ case.

Below, a simple problem is used to visualize the behavior of the
ESLA and compare it with the HLS, and the Rprop algorithm. The
energy landscape of Figure~\ref{fig:single neuron} has a global
minimum and two local minima. Figure~\ref{fig:single neuron} shows
that under the same initial conditions, both of the ESLA and the
HLS escape the saddle point and the valley that leads to a local
minimum, while the ESLA converges faster than HLS with fewer
oscillations(Figure~\ref{fig:single neuron}, left), and the Rprop
algorithm converges to the local minimizer (Figure~\ref{fig:single
neuron}, right).

\vspace*{-0.15cm}
\begin{figure}[htp]
\begin{center}
\includegraphics[width=4.5cm,height=4.2cm]{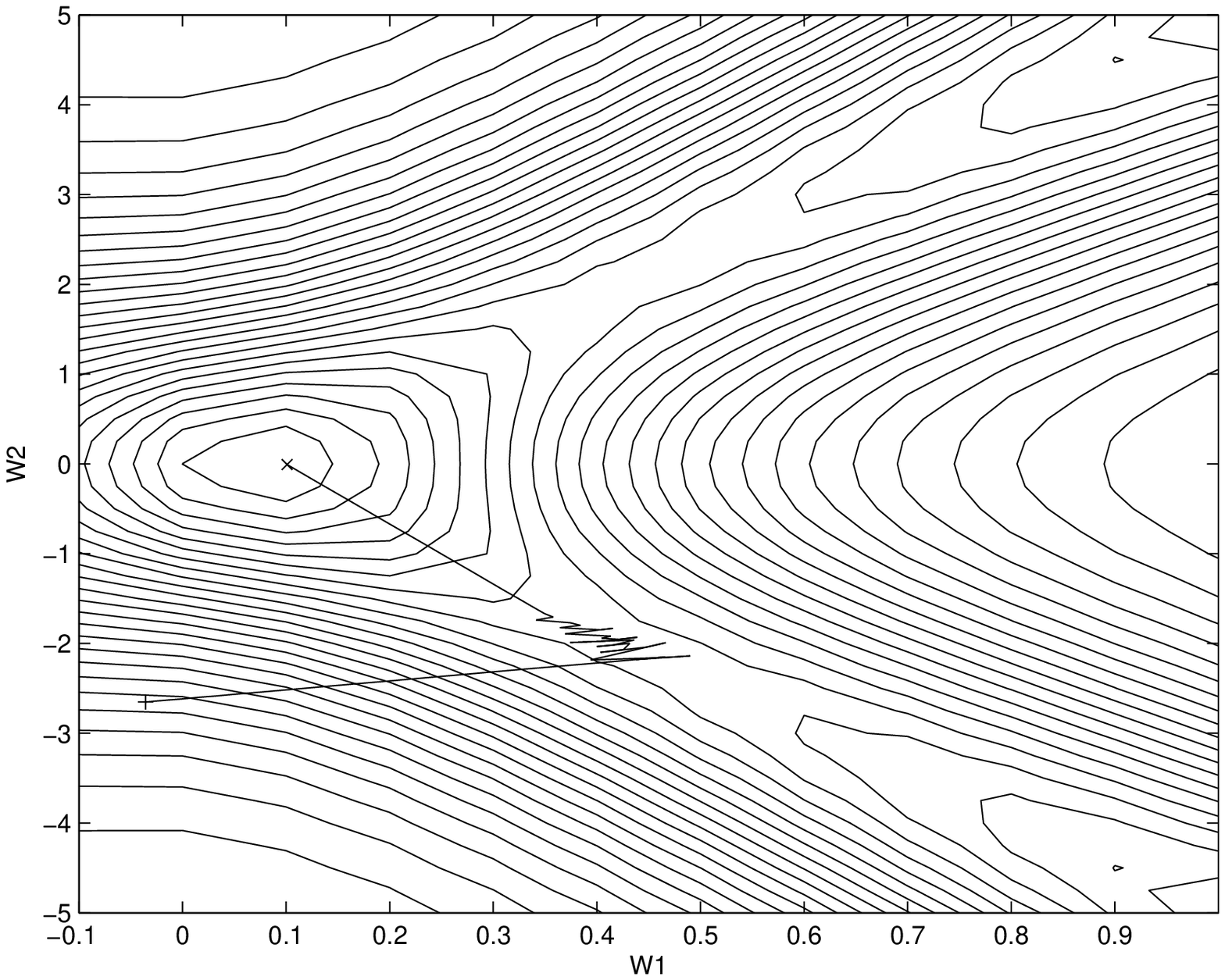}
\includegraphics[width=4.5cm,height=4.2cm]{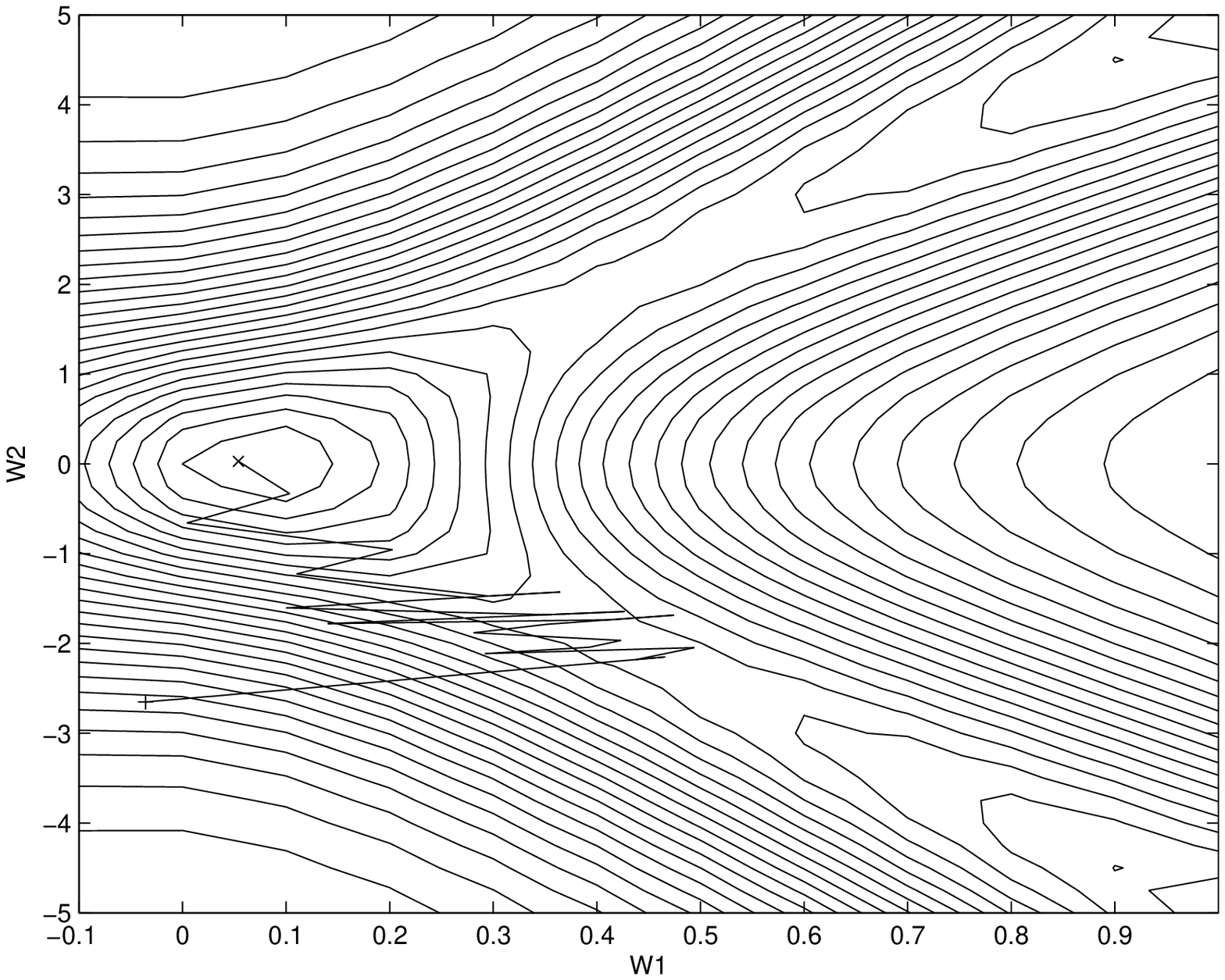}
\includegraphics[width=4.5cm,height=4.2cm]{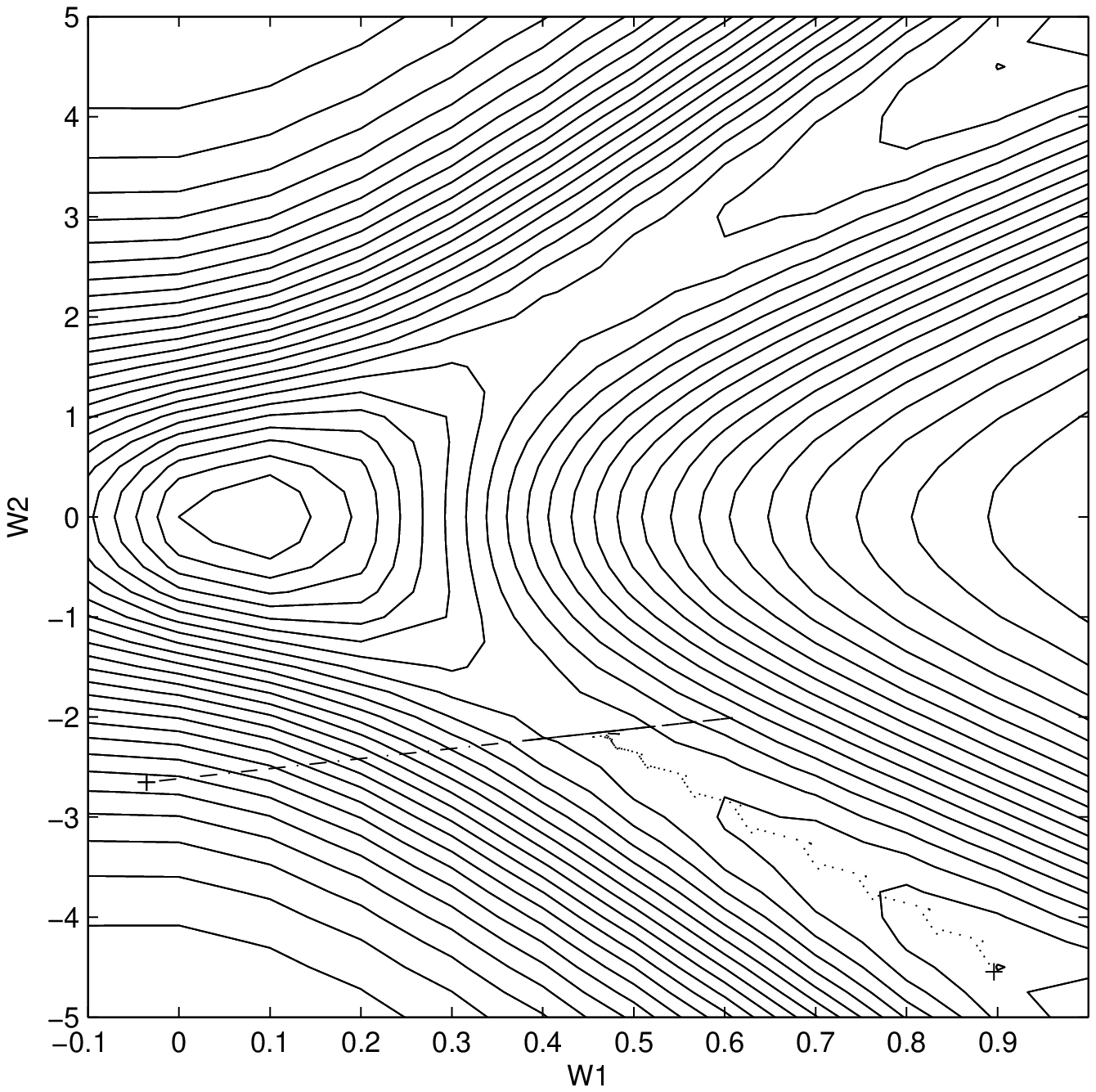}
\vspace*{-0.5truecm} \caption{Weights trajectories of the Evolving
Stochastic Learning Algorithm--ESLA (left), the Hybrid Learning
Scheme--HLS (center), and the Rprop (right).} \label{fig:single
neuron}
\end{center}
\end{figure}
\vspace*{-0.75cm}
\section{Experimental study}
\vspace*{-0.25cm}We have evaluated the performance of the ESLA and
compared it with the Rprop, and the HLS algorithms. The
statistical significance of the results has been analyzed using
the Wilcoxon test~\cite{SnedecorC89}. This is a nonparametric
method that is considered an alternative to the paired $t$--test.
All statements in the tables reported below, refer to a
significance level of $0.05$. Statistically significant cases are
marked with $(+)$, while $(-)$ shows the cases that don't satisfy
the significance level. Moreover, the following terms are used:
$Epochs$ is the number of iterations to converge to the error
target; $Convergence$ denotes the success of convergence to the
error target within $2000$ iterations; $Generalization$ is the
percentage of correctly classified test examples. Finally, for all
the the problems we have set the initial temperature to $T=2$ for
training using the ESLA. By keeping constant the initial
temperature we found the optimal value for the Tsallis entropic
index $q$. The parameters of the HLS were set to the same values
as in the ESLA for all experiments in an attempt to test the
robustness of the method in different types of problems: the
temperature is equal to the initial temperature $T=2$, and the $q$
is set in different values depending on the problem, (i.e. in
cancer $T=2$ and $q=1.7$, while in diabetes is $q=1.6$). Below, we
report results from 300 independent trials. These 300 random
weight initializations have been the same for the three learning
algorithms. \vspace*{-0.5cm}
\subsection{Benchmarks from the UCI Repository}
\begin{figure}[htp]
\vspace*{-0.25truecm}
\begin{center}
\includegraphics[width=4cm,height=3.7cm]{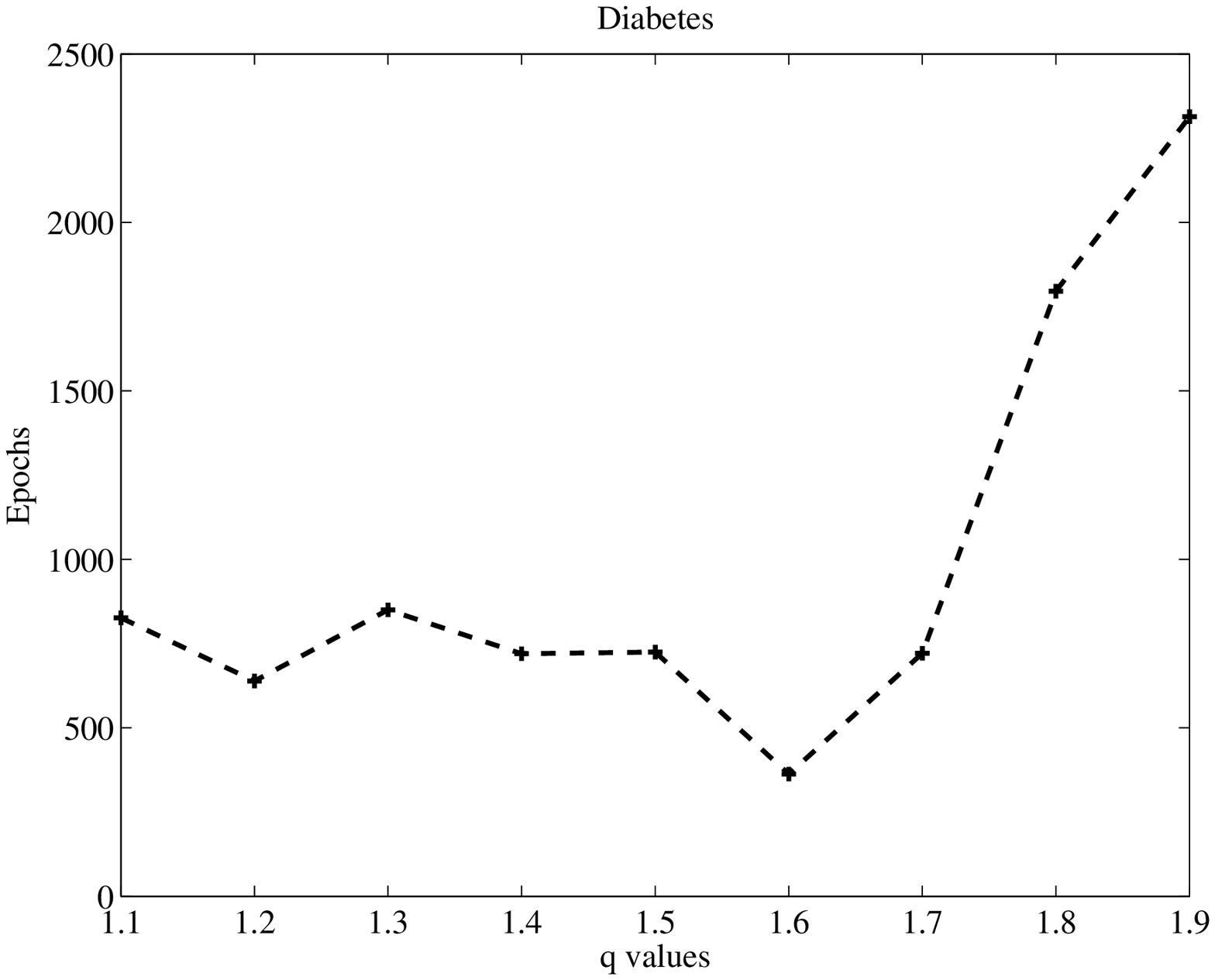}
\includegraphics[width=4cm,height=3.7cm]{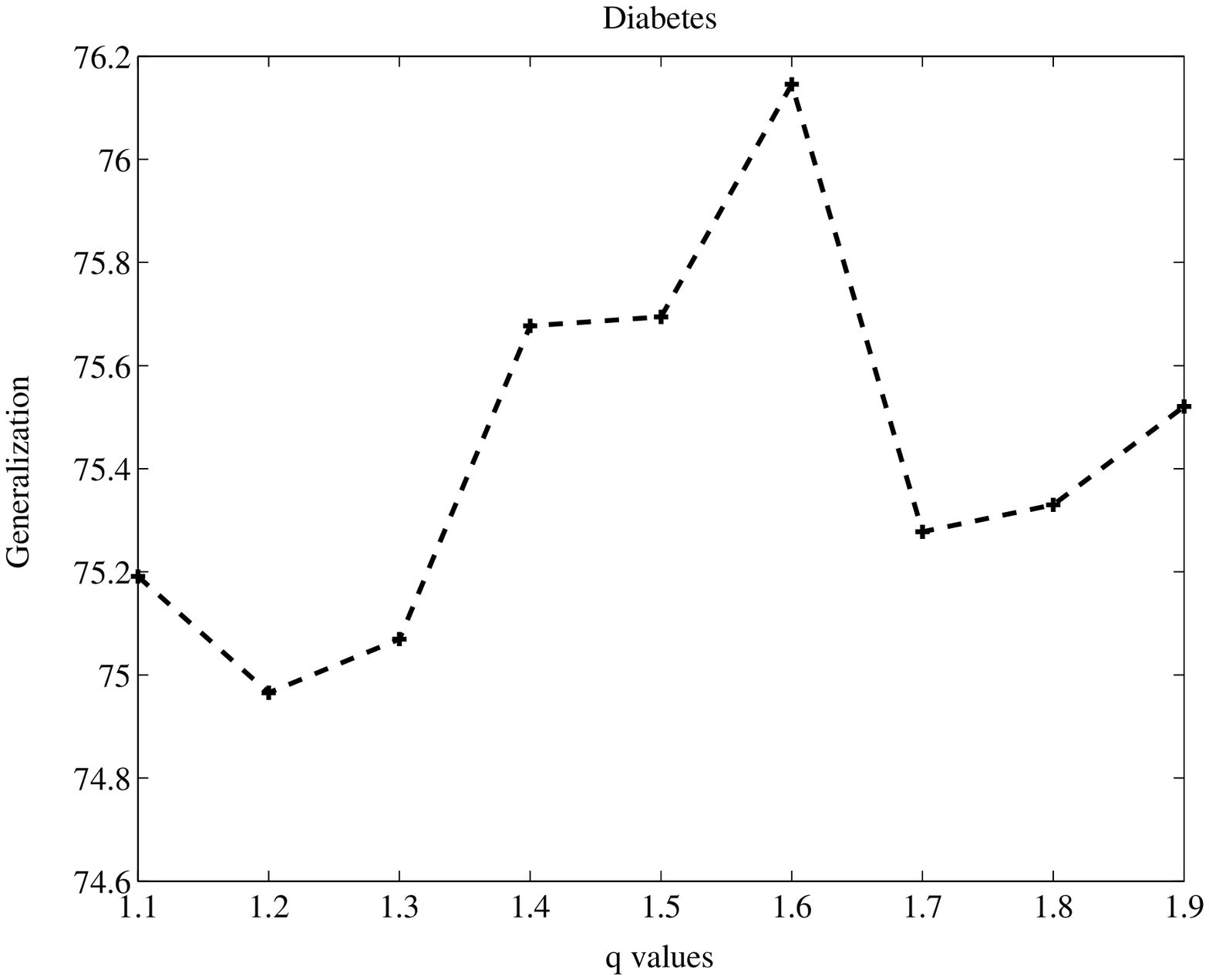}
\includegraphics[width=4cm,height=3.7cm]{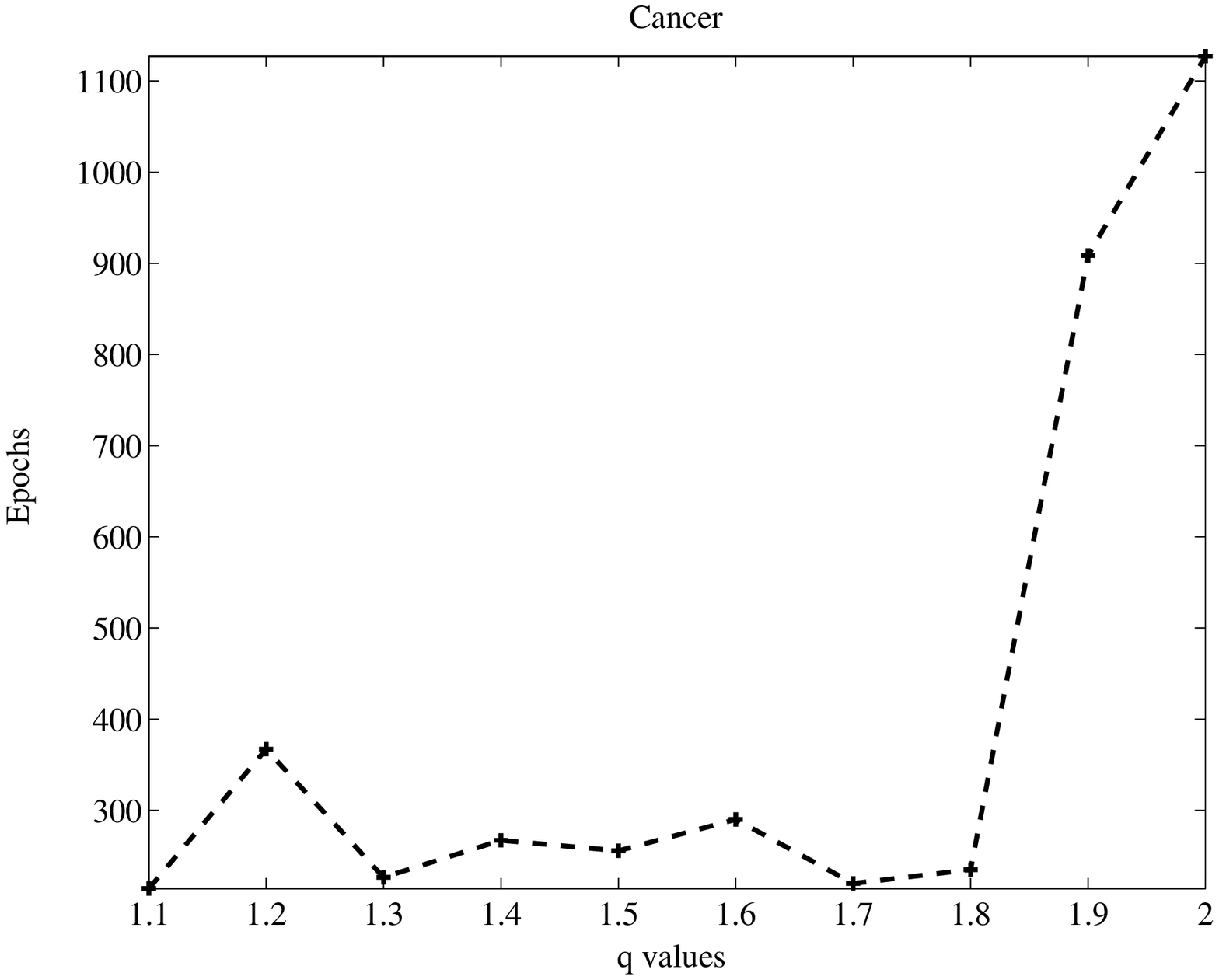}
\includegraphics[width=4cm,height=3.8cm]{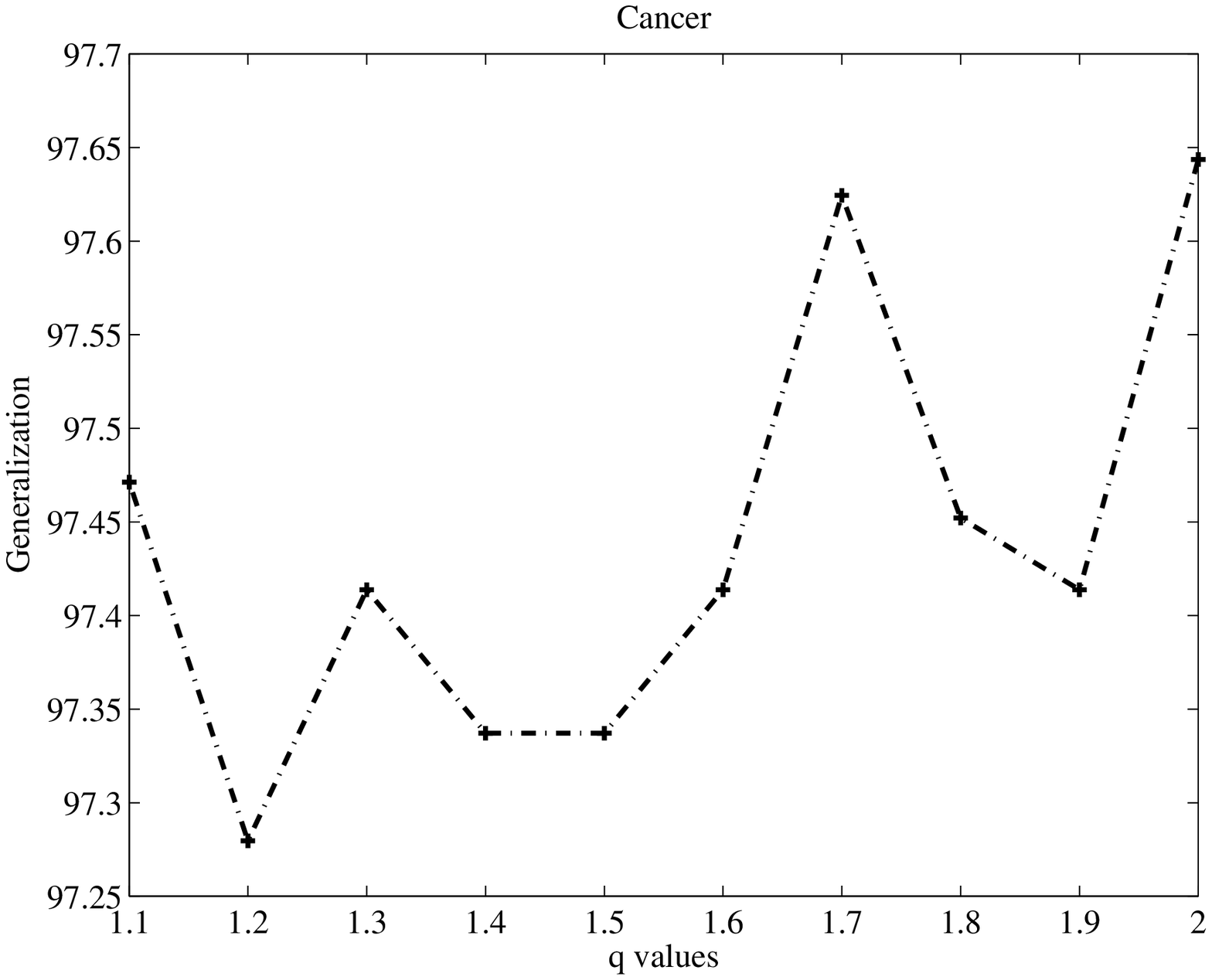}
\vspace*{-0.5truecm}\caption{Optimal \textit{q} based on Epochs,
and Generalization for the diabetes (two left plots), and cancer
problems.} \label{ESLA_diabetes cancer_OptimalQ}
\end{center}
\vspace*{-0.5truecm}
\end{figure}

\vspace*{-0.25cm} The data sets for the cancer1, diabetes1,
thyroid1 problems were used as supplied on the PROBEN1 website.
PROBEN1 provides explicit instructions for creating training and
testing sets and choosing network architectures for many
problems~\cite{Prechelt94}. The partitioning is 50\% of the full
data is used as training set, then the next 25\% of the dataset is
used as validation set, and the remaining 25\% as testing set. The
{\em diabetes1} benchmark is a real-world classification task
which concerns deciding when a Pima Indian individual is diabetes
positive or not~\cite{MurphyA94,Prechelt94}. The Proben1
collection suggests a 8--2--2--2 FNN. The termination criterion is
$E \leq 0.14$ within $2000$ iterations. In order to find the best
value for the initial temperature and the tsallis entropic index
$\textit{q}$, we performed 30 different runs.
Figure~\ref{ESLA_diabetes cancer_OptimalQ} shows the ESLA's
performance for an initial temperature $T=2$ and different $q$
values. Judging from the Figure~\ref{ESLA_diabetes
cancer_OptimalQ} the best value for $\textit{q}=1.6$, and $T=2$.
\vspace*{-0.15cm}
Table~\ref{table:Diabetes Cancer ESLA} shows that the Rprop
algorithm converges many times in local minima. The new stochastic
learning algorithm overcomes this problem in most of the cases.
The cooling procedure seems to have a positive impact on the
learning speed of the algorithm.
\begin{table}[htb]
\vspace*{-0.5cm}
\begin{center}
\begin{scriptsize}
\caption{Comparison of algorithms performance in the Diabetes and
Cancer problems for the converged runs} \label{table:Diabetes
Cancer ESLA}
\newcommand{\m}{\hphantom{$-$}}
\newcommand{\cc}[1]{\multicolumn{1}{c}{#1}}
\renewcommand{\tabcolsep}{0.7pc} 
\renewcommand{\arraystretch}{1.0} 
\begin{tabular}{@{}lllllll}
\hline
          &  \textbf{Diabetes}  &                       &                   & \textbf{Cancer} &         &                           \\
Algorithm &    $Epochs$         & $Generalization$      & $Convergence$     & $Epochs$         & $Generalization$  & $Convergence$ \\
\hline
Rprop     &700$\;(+)$           & 75.2 (\%)$\;(+)$      & 86 (\%)$\;(+)$     &287 $\;(+)$      & 97.2(\%) $\;(-)$  & 94(\%) $\;(+)$  \\
HLS       &570$\;(+)$           & 75.8 (\%)$\;(+)$      & 94 (\%)$\;(-)$     &230 $\;(+)$      & 97.4(\%) $\;(-)$  & 96(\%) $\;(+)$  \\
ESLA      &480                  & 76.2  (\%)            & 95  (\%)           &195              & 97.4(\%)          & 99(\%) \\
\hline
\end{tabular}
\end{scriptsize}
\end{center}
\vspace*{-0.5cm}
\end{table}
\noindent %
\vspace*{-0.15cm}
The second benchmark is the {\em breast cancer diagnosis} problem
which classifies a tumor as benign or malignant based on 9
features~\cite{MurphyA94,Prechelt94}. We have used an FNN with
9--4--2--2 nodes, as suggested in~\cite{Prechelt94}, and a
termination criterion of $E \leq 0.02$. Figure~\ref{ESLA_diabetes
cancer_OptimalQ} shows the best values of these two important
training parameters. As we can observe from this figure, a value
of the $\textit{q}=1.7$ gives the best results in terms of both
learning speed and generalization. The comparative results are
presented in Table~\ref{table:Diabetes Cancer ESLA}.

The third benchmark problem is the {\em thyroid1}, which is not a
permutation of the original data, but retains the original order
instead \cite{MurphyA94,Prechelt94}. The data set consists of 3600
patterns. The termination criterion is $E \leq 0.0036$. The
Tsallis entropic index $q$ in this problem is again $q=1.7$. The
experimental results that we obtained are presented in
Table~\ref{table:Thyroid Yeast ESLA}. \vspace*{-0.15cm}
\begin{figure}[htp]
\begin{center}
\includegraphics[width=4cm,height=3.7cm]{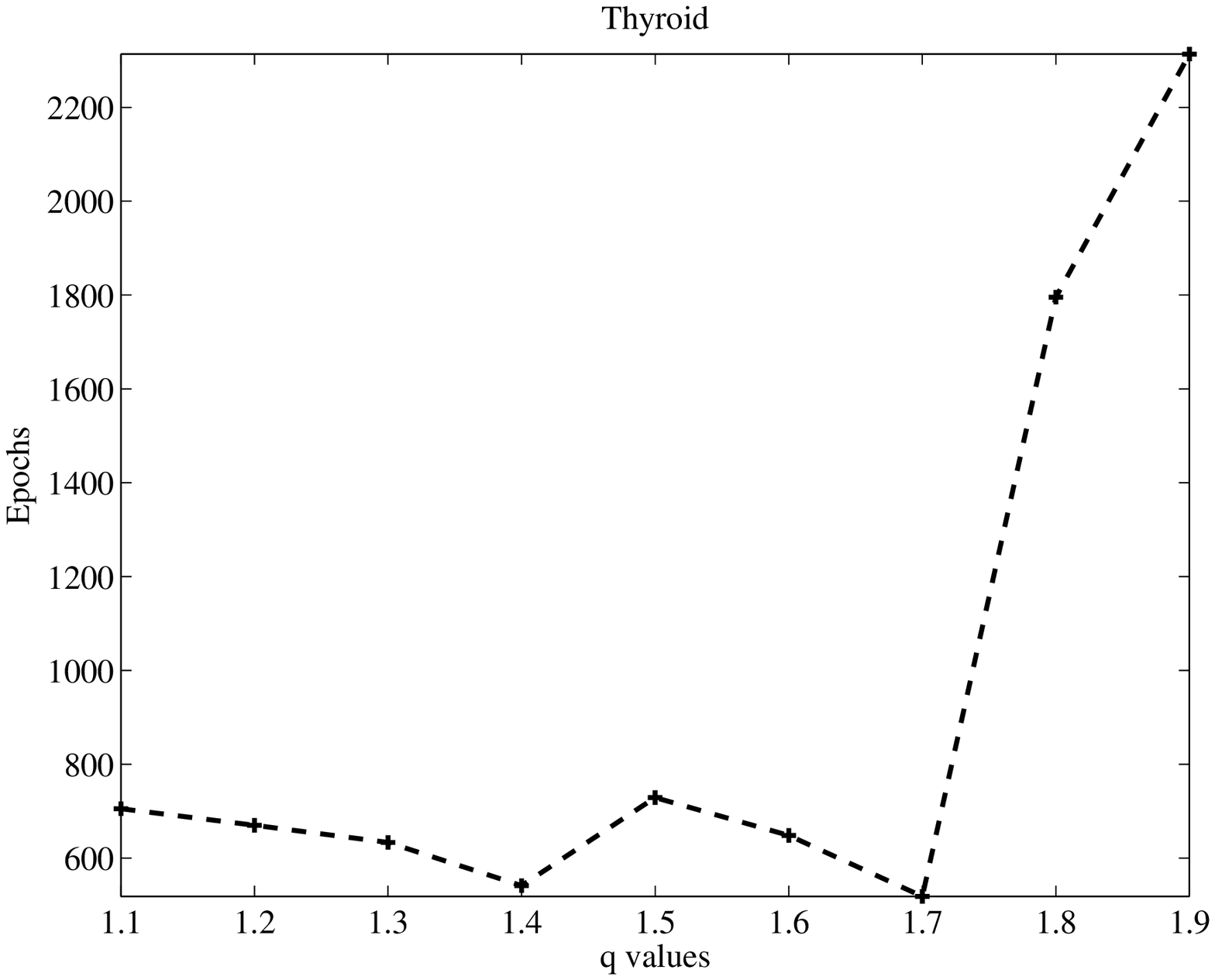}
\includegraphics[width=4cm,height=3.7cm]{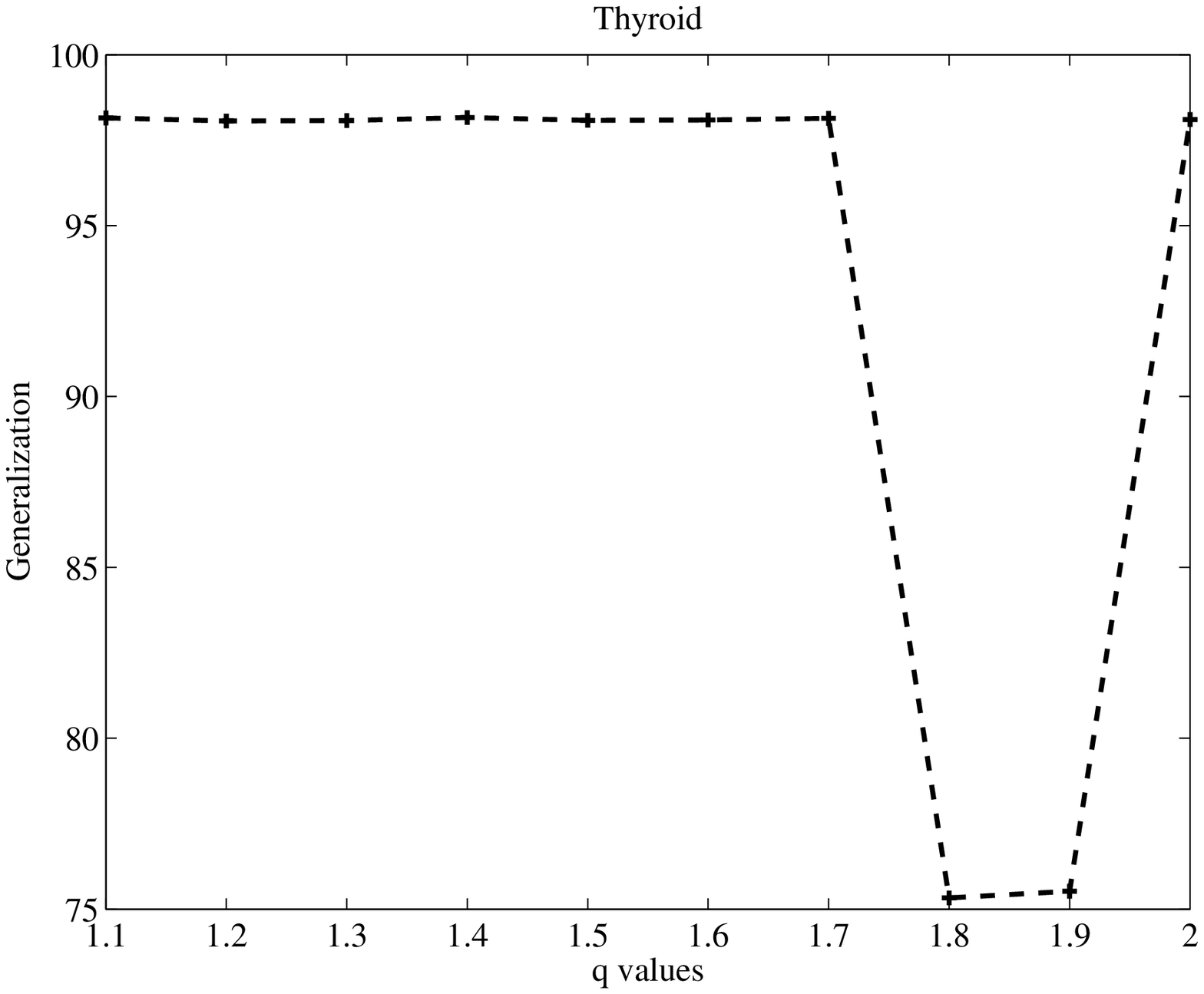}
\includegraphics[width=4cm,height=3.7cm]{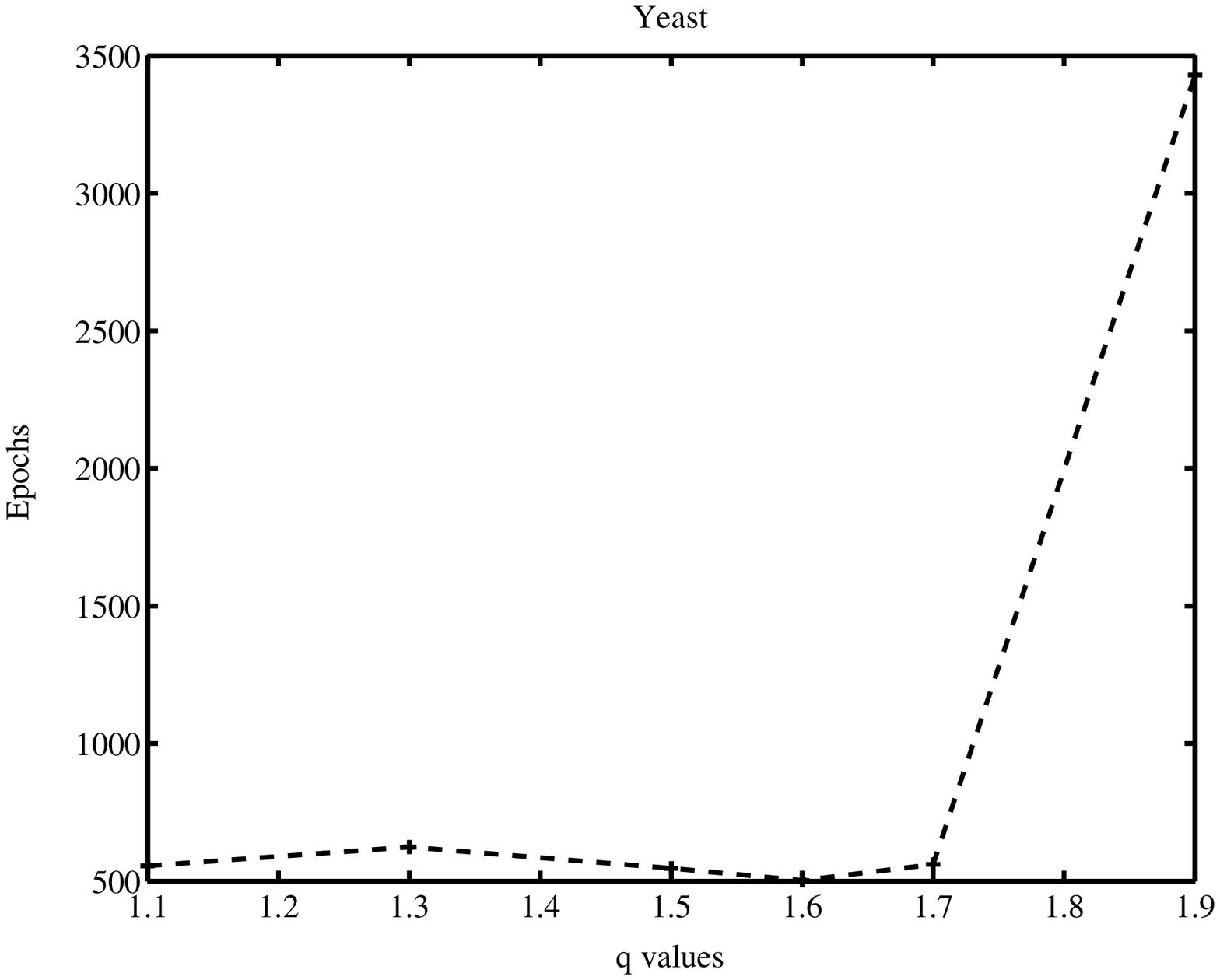}
\includegraphics[width=4cm,height=3.7cm]{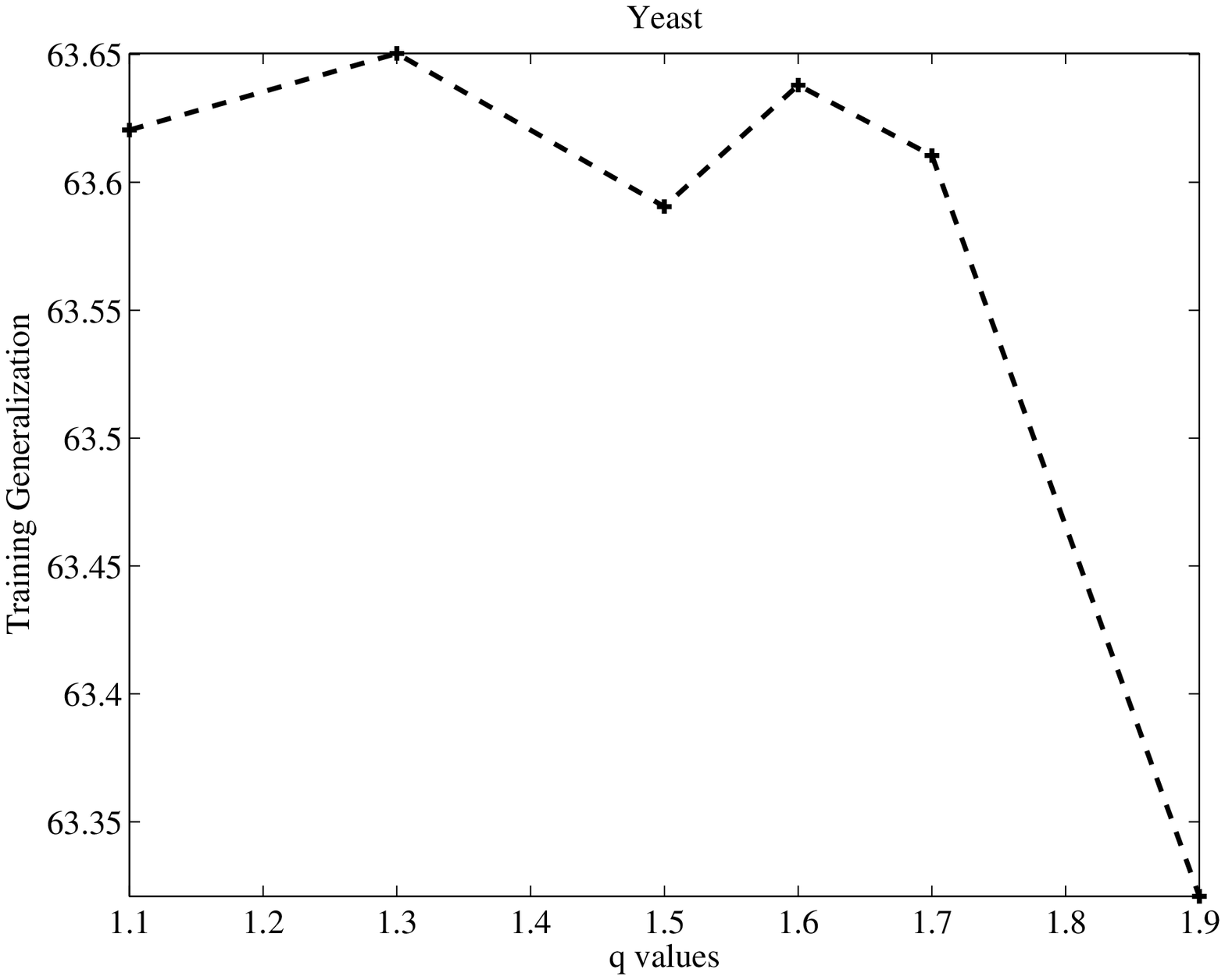}
\caption{Optimal \textit{q} based on Epochs, and Generalization
for the thyroid (two left plots), and Yeast problems.}
\label{ESLA_thyroid Yeast_OptimalQ}
\end{center}
\end{figure}
\begin{table}[htb]
\vspace*{-0.25cm}
\begin{center}
\begin{scriptsize}
\caption{Comparison of algorithms performance in the Thyroid and
Yeast problems for the converged runs} \label{table:Thyroid Yeast
ESLA}
\newcommand{\m}{\hphantom{$-$}}
\newcommand{\cc}[1]{\multicolumn{1}{c}{#1}}
\renewcommand{\tabcolsep}{0.7pc} 
\renewcommand{\arraystretch}{1.0} 
\begin{tabular}{@{}lllllll}
\hline
          &  \textbf{Thyroid}  &                     &                   & \textbf{Yeast} &         &                           \\
Algorithm &    $Epochs$        & $Generalization$   & $Convergence$     & $Epochs$         & $Generalization$  & $Convergence$ \\
\hline
Rprop           &780$\;(+)$    & 98.2 (\%)$\;(-)$  & 81.3 (\%)$\;(+)$   &930$\;(+)$     & 61.6 (\%)$\;(-)$  & 98 (\%)$\;(-)$  \\
HLS             &590$\;(+)$    & 98.1 (\%)$\;(-)$  & 94.0 (\%)$\;(-)$   &590$\;(+)$     & 61.4 (\%)$\;(-)$  & 100 (\%)$\;(-)$  \\
ESLA            &500           & 98.0  (\%)        & 95.3  (\%)         &490            & 61.5  (\%)        & 100  (\%)         \\
\hline
\end{tabular}
\end{scriptsize}
\end{center}
\end{table}
\vspace*{-0.25cm}
\vspace*{-0.5cm}
\subsection{Prediction of Localisation sites of the Yeast Proteins}
\vspace*{-0.25cm}The study of protein localization is considered
very useful in the post-genomics and proteomics era, as it
provides information about each protein that is complementary to
the protein sequence and structure data~\cite{Boland99}. One of
the most thoroughly studied single--cell organisms is the
eukaryote Saccharomyces cerevisiae, also called Yeast. It has
rapid growth rate and very simple nutritional
requirements~\cite{Lodish03}. The Yeast dataset is 1484 proteins
labeled according to 10 sites~\cite{Horton97}. Yeast proteins are
organized as in~\cite{MurphyA94}. The most suitable architecture
for this problem, as suggested by~\cite{AnastasiadisM2003IDA}, is
an 8-16-10 FNN architecture. A termination criterion of $E \leq
0.05$ within $2000$ iterations ($Epochs$) is used. The evaluation
method that we have employed to estimate the accuracy of the
methods was a $10$-fold cross validation following the guidelines
of~\cite{AnastasiadisM2003IDA,Horton97}. The proportion of the
number of the patterns for all the classes is equal in each
partition, as this procedure provides more accurate results than a
plain cross validation does~\cite{Kohavi95}.
Figure~\ref{ESLA_thyroid Yeast_OptimalQ} gives an overview of the
experiments conducted in order to choose the best value of $q$ for
this problem. A value of $q=1.6$ was applied as this gave the best
results in terms of learning speed and generalization.
Table~\ref{table:Thyroid Yeast ESLA} shows the experimental
results for this difficult problem. \vspace*{-0.5cm}
\subsection{Boolean function approximation problems}
\vspace*{-0.25cm}Another set of experiments has been conducted to
empirically evaluate the performance of the new method in a
well--studied class of boolean function approximation problems
that exhibit strong local minima~\cite{Blum89}. This class
includes the XOR problem, and the parity--$3$ problem, which is
considered as classic
benchmarks~\cite{AnastasiadisM2004a,TreadgoldG98}. The adopted
architectures for the XOR problem is a 2--2--1, and the error
target was set to $E \leq 10^{-5}$. A 3--3--1 architecture was
used for the parity--$3$ problem. The error target for parity-$3$
problem was set to $E \leq 5\times 10^{-5}$. The activation
function for this problem is the tansig function. These target
values are considered low enough to guarantee convergence to a
``global'' solution. \vspace*{-0.15cm}

By applying the same procedure as before, the best $q$ entropic
index value for the XOR problem is $q=2.1$, and for the parity $3$
problem is $q=1.1$ with initial temperature $T=2$.
Table~\ref{table: XOR and Parity 3 ESLA} shows that the ESLA
outperforms in convergence speed. The HLS achieves the best
Convergence success on XOR problem. However, the ESLA has better
convergence performance compared to Rprop.
\begin{table}[h!]
\vspace*{-0.25cm}
\begin{center}
\begin{scriptsize}
\caption{Comparison of algorithms performance in the XOR and
Parity $3$ problems for the converged runs} \label{table: XOR and
Parity 3 ESLA}
\newcommand{\m}{\hphantom{$-$}}
\newcommand{\cc}[1]{\multicolumn{1}{c}{#1}}
\renewcommand{\tabcolsep}{0.7pc} 
\renewcommand{\arraystretch}{1.0} 
\begin{tabular}{@{}lllllll}
\hline
          &  \textbf{XOR}  &                     &                   & \textbf{Parity $3$} &         &                           \\
Algorithm &    $Epochs$        & $Generalization$   & $Convergence$     & $Epochs$         & $Generalization$  & $Convergence$ \\
\hline
Rprop           &120$\;(+)$  & 100 (\%)$\;(-)$    & 59 (\%)$\;(+)$   &877$\;(+)$     & 100 (\%)$\;(-)$  & 74 (\%)$\;(+)$  \\
HLS             &80$\;(+)$    & 100 (\%)$\;(-)$   & 68 (\%)$\;(-)$   &430$\;(+)$     & 100 (\%)$\;(-)$  & 78 (\%)$\;(+)$  \\
ESLA            &70           & 100 (\%)          & 64  (\%)         &390            & 100  (\%)        & 81  (\%)         \\
\hline
\end{tabular}
\end{scriptsize}
\end{center}
\end{table}
\vspace*{-0.25cm}
\vspace*{-0.5cm}
\section{Discussion and Concluding Remarks}
\vspace*{-0.25cm}A recently introduced training algorithm, the
hybrid learning scheme-HLS achieves generally very good and
reliable performance, and improved learning speed compared to the
Rprop algorithm. In this paper, we proposed a new evolving
stochastic learning scheme, which constitutes an efficient
improvement of the HLS algorithm that is built on a theoretical
basis. The ESLA combines deterministic and stochastic search by
employing a different adaptive stepsize for each weight, and a
form of noise that is characterized by the nonextensive entropic
index $q$. An adaptive formula that introduces a relationship
between the $T$ and $q$ was applied. Our experimental study showed
that there is a range of $q$ values ($1.1<q<2.3$) that gives good
performance for the new learning scheme.

In previous tables the results are based only on the converged
runs. Therefore, we don't have the actual performance description
of the tested algorithms (\emph{i.e.} in thyroid problem the Rprop
algorithm achieves the best mean generalization success. However,
its convergence success is the worst within the tested algorithms.
Therefore, the convergence results present the Rprop's
generalization for the $0.813\cdot 300=244$ runs out of 300, while
the mean generalization success of ESLA is based on $0.953\cdot
300= 286$ runs out of 300). In this case it is better to have
results for more runs (\emph{i.e.} patients) although the
generalization success is slightly worse. In order to have better
view of the overall performance of the tested algorithms, we
introduce the parameter \emph{Performance}, which is defined as
follows:
$\emph{Performance}=\frac{(Convergence)\times{(Generalisation)}}{100}$.
Thus, Table~\ref{Performance ESLA} gives a summary of our results
from this perspective for all the tested algorithms.
\begin{table}[htp]
\vspace*{-0.25cm}
\begin{center}
\begin{scriptsize}
\caption{Summary of the results in terms of the algorithms'
Performance} \label{Performance ESLA}
\newcommand{\m}{\hphantom{$-$}}
\newcommand{\cc}[1]{\multicolumn{1}{c}{#1}}
\renewcommand{\tabcolsep}{0.7pc} 
\renewcommand{\arraystretch}{1.0} 
\begin{tabular}{@{}lllllllll}
\hline
\textbf{Performance}        &     &  \textbf{Algorithms}   &   &  \\
\hline

Problems                       &Rprop (\% )    & HLS (\%)        & ESLA (\%)    \\
\hline \textbf{Diabetes}&
                               64.7            &71.2             & 72.4         \\
\textbf{Cancer}&
                               91.4           &93.5               & 96.4         \\
\textbf{Thyroid}  &
                               79.8           &92.3            & 93.6            \\
\textbf{Yeast}&
                               60.3           &61.4           & 61.5           \\
\textbf{XOR}&
                               59.0       &   68.0              & 64.0           \\
\textbf{Parity--$3$}&
                              74.0             & 78.0          & 81.0          \\
\hline
\end{tabular}
\end{scriptsize}
\end{center}
\end{table}
\vspace*{-0.25cm}

Further testing is of course necessary to fully explore the
advantages and identify possible limitations of this cooling
evolving scheme. Moreover, exhaustive testing of the new method in
other classes of problems will be done. We will also investigate
the performance of ESLA in a restarting mode. Finally, we are
going to explore further the properties of Tsallis entropy into
Optimization methods in Artificial Intelligence applications.
\vspace*{-0.5cm}
\section{Acknowledgements}
\vspace*{-0.25cm}Aristoklis Anastasiadis would like to thank Dr.
G. Kaniadakis and would also like to address special thanks to
Prof. Constantino Tsallis for very helpful discussions related to
this work, during his stay as research visitor at the Santa Fe
Institute. \vspace*{-0.25cm}

\end{document}